\documentclass[letterpaper]{article} % DO NOT CHANGE THIS
\usepackage{aaai2026} 
\usepackage{times}  % DO NOT CHANGE THIS
\usepackage{helvet}  % DO NOT CHANGE THIS
\usepackage{courier}  % DO NOT CHANGE THIS
\usepackage[hyphens]{url}  % DO NOT CHANGE THIS
\usepackage{graphicx} % DO NOT CHANGE THIS
\usepackage{natbib}  % DO NOT CHANGE THIS AND DO NOT ADD ANY OPTIONS TO IT
\usepackage{caption} % DO NOT CHANGE THIS AND DO NOT ADD ANY OPTIONS TO IT
\usepackage{algorithm}
\usepackage{algorithmic}

\usepackage{newfloat}
\usepackage{listings}
\DeclareCaptionStyle{ruled}{labelfont=normalfont,labelsep=colon,strut=off} % DO NOT CHANGE THIS
\floatstyle{ruled}
\newfloat{listing}{tb}{lst}{}
\floatname{listing}{Listing}
\usepackage{times}
\usepackage{latexsym}
\usepackage{microtype}
\usepackage{graphicx}
\usepackage{subfigure}
\usepackage{booktabs} % for professional tables
\usepackage{amsmath}
\usepackage{amssymb}
\usepackage{mathtools}
\usepackage{amsthm}

\makeatletter
\DeclareRobustCommand\onedot{\futurelet\@let@token\@onedot}
\def\@onedot{\ifx\@let@token.\else.\null\fi\xspace}

\def\ie{\emph{i.e}\onedot}

\makeatother

\usepackage{multirow}

\usepackage[T1]{fontenc}

\usepackage{colortbl}
\definecolor{mygray}{gray}{.9}
\usepackage{xspace}

\title{Text-Driven Emotionally Continuous Talking Face Generation}
\author{\small
Hao Yang$^{1}$,
Yanyan Zhao$^{1}$,
Kewei Zhao$^{1}$,
Hongbo Zhang$^{1}$,
Tian Zheng$^{1}$,
Yusheng Liu$^{1}$,\\
Xing Fu$^{1}$,
Bichen Wang$^{1}$,
Bing Qin$^{1}$,
Hao He$^{2}$,
Zhen Wu$^{2}$,
Xuda Zhi$^{2}$,
Yongbo Huang$^{2}$ \\
$^{1}$Harbin Institute of Technology \quad
$^{2}$SERES \\
{\tt\small \{hyang, yyzhao, kwzhao, hbzhang, tzheng, ysliu, xfu, bcwang, qinb\}@ir.hit.edu.cn}
}
\usepackage{bibentry}
\begin{document}

\maketitle

\begin{abstract}
Talking Face Generation (TFG) strives to create realistic and emotionally expressive digital faces. 
While previous TFG works have mastered the creation of naturalistic facial movements, they typically express a fixed target emotion in synthetic videos and lack the ability to exhibit continuously changing and natural expressions like humans do when conveying information.
To synthesize realistic videos, we propose a novel task called \textbf{E}motionally \textbf{C}ontinuous \textbf{T}alking \textbf{F}ace \textbf{G}eneration (EC-TFG), which takes a text segment and an emotion description with varying emotions as driving data, aiming to generate a video where the person speaks the text while reflecting the emotional changes within the description.
Alongside this, we introduce a customized model, \ie, 
\textbf{T}emporal-\textbf{I}ntensive \textbf{E}motion Modulated \textbf{T}alking \textbf{F}ace \textbf{G}eneration (TIE-TFG), which innovatively manages dynamic emotional variations by employing Temporal-Intensive Emotion Fluctuation Modeling, allowing to provide emotion variation sequences corresponding to the input text to drive continuous facial expression changes in synthesized videos.
Extensive evaluations demonstrate our method's exceptional ability to produce smooth emotion transitions and uphold high-quality visuals and motion authenticity across diverse emotional states. Further visualization can be found at Supplementary Material.
\end{abstract}

\section{Introduction}
\label{sec:intro}

\begin{figure}

\begin{center}

\includegraphics[width=7.5cm]{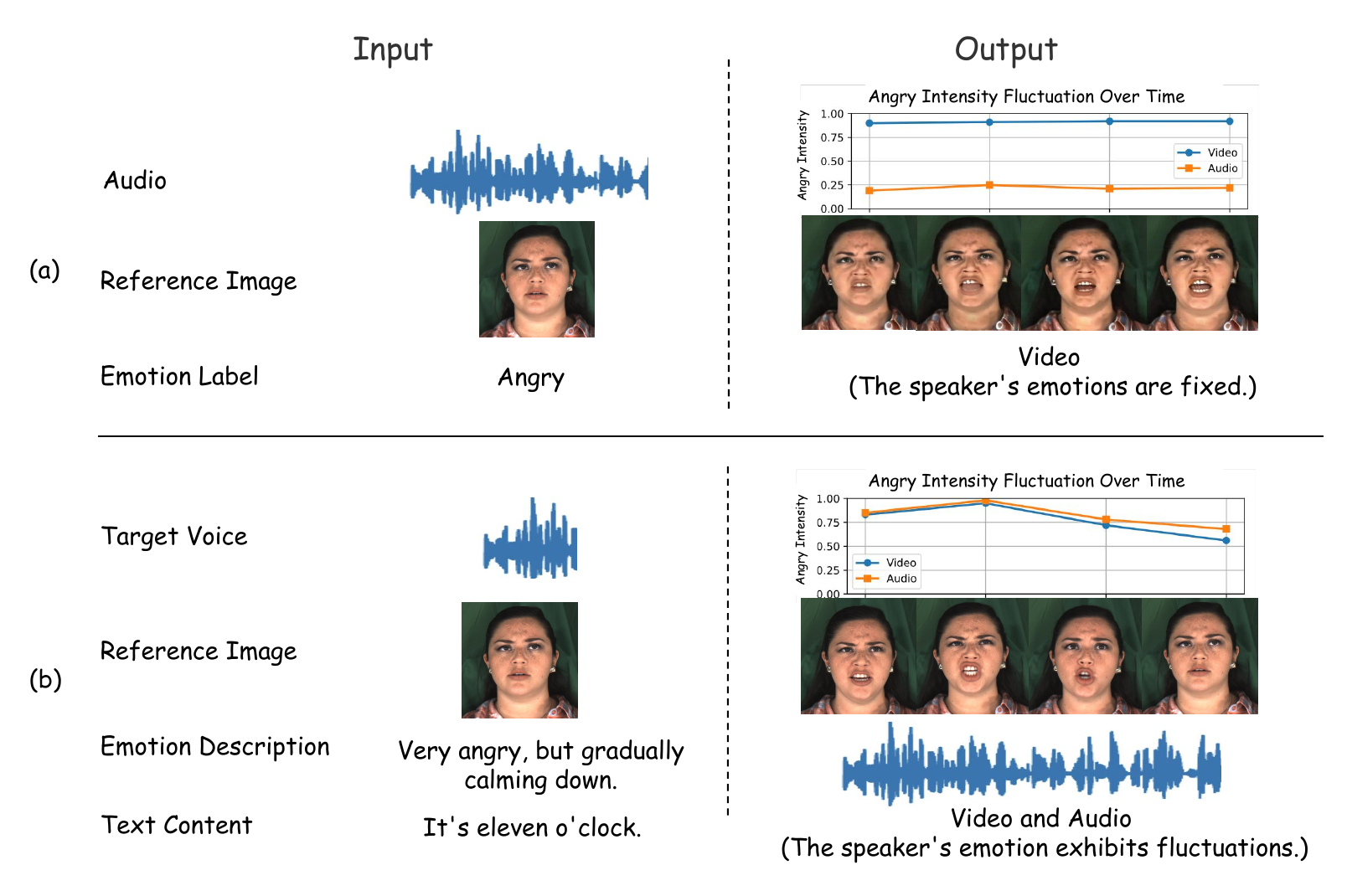} 

\caption{
Comparison between (a) the traditional audio-driven emotional TFG task and (b) the proposed EC-TFG task.
}
\label{fig:task}
\end{center}

\end{figure}

Talking Face Generation (TFG) aims to create realistic videos by utilizing reference images and driving data, such as audio or text. 
This task is challenging and holds significant practical value across diverse domains like film production. 
% and virtual reality.
Recently, the development of TFG has accelerated~\cite{prajwal2020lip,he2023gaia,ma2023dreamtalk,shen2023difftalk,stypulkowski2024diffused,sun2023vividtalk,corona2024vlogger,liu2024anitalker,tian2024emo,wei2024aniportrait,xu2024vasa,cheng2024dawn,chatziagapi2024talkinnerf,hogue2024diffted,su2024audio}, leading to significant quality improvements of synthesized videos. 
While existing TFG works are suitable for neutral emotions, diverse and dynamic emotions are vital for intent, sparking interest in editable emotional TFG, which takes a fixed audio signal and a target emotion as input to synthesize an emotion-controllable video.

However, existing emotional TFG works~\cite{ji2021audio,eskimez2021speech,liang2022expressive,zhang2024emotalker,sinha2022emotion} are limited to audio-driven, inherently struggle to synthesize videos where emotional changes are continuous and natural, with seamless coordination between audio and visual perception, as humans express continuous emotions in the real world.
Consequently, altering the input emotion may result in synthesized video content that conflicts with the emotions in the original audio signal, as shown in Figure 1(a). 
In addition, existing works still focus on creating videos with a fixed emotion, which does not align with the dynamic and ever-changing emotional expressions in human behavior.
On the other hand, text-driven TFG provides the text to be spoken in the synthesized video, which offers the potential to generate visually and audibly coordinated videos when changing the target emotion.
Current text-driven TFG researches~\cite{zhang2022text2video,ye2023ada,wang2023text,mitsui2023uniflg,jang2024faces} remain focused on combining TTS and TFG processes, overlooking the significant potential for emotion editing.

To move towards synthesizing talking face that approximates the real video, we explore a novel task for the first time: \textbf{E}motionally \textbf{C}ontinuous and \textbf{N}atural \textbf{T}alking \textbf{F}ace \textbf{G}eneration (EC-TFG).
As shown in Figure 1(b), EC-TFG takes as input an emotion description with varying emotions instead of a fixed emotion label and the corresponding text to be spoken, with the goal of driving the generation of talking faces that coherently and smoothly reflect the emotion description, in coordination with audio-visual information.

To achieve the goals of EC-TFG, we propose a novel customized method, \ie, \textbf{T}emporal-Intensive  \textbf{E}motion \textbf{M}odulated \textbf{T}alking \textbf{F}ace \textbf{G}eneration (TIE-TFG), which guides the diffusion model to synthesize the emotionally continuous and natural video by transforming the input emotion description into a fine-grained emotion sequence.
Specifically, TIE-TFG first utilizes a pre-trained Text-To-Speech model to process the voice reference, the text to be spoken, and the emotion description, thereby synthesizing an audio signal that incorporates the desired emotional variations.
Subsequently, the synthesized audio signal is processed by an audio encoder and, together with the textual features, is fed into a newly introduced Temporal-Intensive Emotion Fluctuation Predictor, which determines the emotion label and intensity corresponding to each word.
Finally, we combine audio features and emotion fluctuation features to decouple lip, face, and head motion information to help generate visual results with emotional fluctuations.

To quantitatively evaluate model performance on EC-TFG, we introduce EC-HDTF, a newly annotated dataset with over 10 hours of emotional videos.
Furthermore, we design a new evaluation metric, Emotional Fluctuation Score, tailored to the goal of EC-TFG,  which assesses whether the emotional fluctuations in the synthesized videos meet the expected outcomes.
We conduct comprehensive comparisons of the newly proposed TIE-TFG with existing methods on EC-HDTF. 
The results reveal that, compared to existing methods, TIE-TFG generates talking faces that more comprehensively and accurately capture the emotions embedded in the descriptions. Additionally, TIE-TFG demonstrates superior control over continuous emotional transitions, showing significant improvements in both the continuity and naturalness of emotional changes.

To summarize, we make the following contributions:
\begin{itemize}
\item Current audio-driven emotional editing in TFG is limited to altering the speaker's facial expressions, without adjusting the emotions conveyed in the audio. To address this, we propose the EC-TFG task that enables synchronized emotional editing for both video and audio.
\item We present the first text-driven talking face generation framework that models emotional fluctuations. Unlike previous methods that control only fixed emotion categories and intensities, our framework models dynamic emotional changes based on text content, allowing for more fine-grained control through emotion description.
\item We are the first to evaluate the capability of TFG models in generating continuous emotional fluctuations. Experiments show that our framework produces more realistic and smoother facial expression changes under emotional control.
\end{itemize}

\section{Related Work}
\subsection{Emotional Talking Face Generation}

Emotionally rich and vivid videos generated by emotional TFG models offer broader applicability than neutral ones, especially in domains such as virtual reality and film production, which have gained increasing attention in recent years. 
Existing emotional TFG models are almost entirely audio-driven, focusing on aligning audio with lip movements while maintaining facial expressions that match one-hot emotion labels. 
For example, customized EVP~\cite{ji2021audio} models generate specific identity-focused expressions by decoupling emotion learning from audio and aligning audio-landmarks. They explore emotion editing by interpolating within the emotional feature space. Recent works have shifted to a one-time setup driven by reference faces for facial expression control. For instance, EAMM~\cite{ji2022eamm} extracts local emotional displacements from additional emotional videos and applies them to reference portraits rendered from audio, achieving expression control for any target subject. EMMN~\cite{tan2023emmn} separates the face into expression and mouth features to globally reconstruct dynamic facial movements. However, the coarse-grained emotion labels limit flexibility in emotion editing. Emotalker~\cite{zhang2024emotalker} designs a method to convert emotion descriptions into corresponding emotion and intensity values, expanding the range of customizable emotional expressions. Nevertheless, these approaches introduce fixed, unchanging emotional features across the sequence, neglecting the temporal fluctuations of emotion according to speech content. In concurrent work, ~\citet{xu2024learning} proposes to model intensity directly from audio in an emotion-agnostic way, achieving frame-level fluctuations in emotional intensity. However, transferring different emotions onto intensity sequences overlooks the relationship between emotion and intensity.

Due to the constant audio input in these models, the generated videos often convey mismatched emotions between the audio and video, making them appear rigid in most cases.
To address these limitations, we propose a text-driven emotional TFG framework that supports fine-grained, customizable emotion control and models emotion intensity fluctuations based on the same driving text across different emotions. Our approach generates more expressive and emotionally coherent talking face videos across diverse reference identities.

\subsection{Text-Driven Talking Face Generation}
While audio-driven talking face generation (TFG) has been extensively studied, text-driven TFG remains in its early stages of development. Current approaches to text-driven TFG can be broadly categorized into two types. The first~\cite{zhang2022text2video,ye2023ada,wang2023text} is a pipeline approach, where synthetic speech is generated from text using a TTS model and then use the output for audio-driven talking face generation. The second~\cite{mitsui2023uniflg,jang2024faces} is an end-to-end approach that combines TTS and TFG tasks by using latent features from the TTS model in place of the TFG audio encoder’s latent features, simultaneously generating both speech and talking face video. Each approach has distinct advantages and limitations~\cite{choi2023reprogramming}: the pipeline approach can leverage large-scale, pre-trained TTS models to generate high-quality audio without additional training but tends to have lower performance and longer inference times. In contrast, end-to-end models provide faster inference but require more intensive training and generally produce less effective joint generation results compared to single-task outputs. In this paper, we adopt a pipeline approach to achieve improved generation quality. By enabling the editing of both audio and video, text-driven TFG inherently provides greater flexibility and controllability.

\section{Method}

\begin{figure*}[t]
  \centering
   \includegraphics[width=15cm]{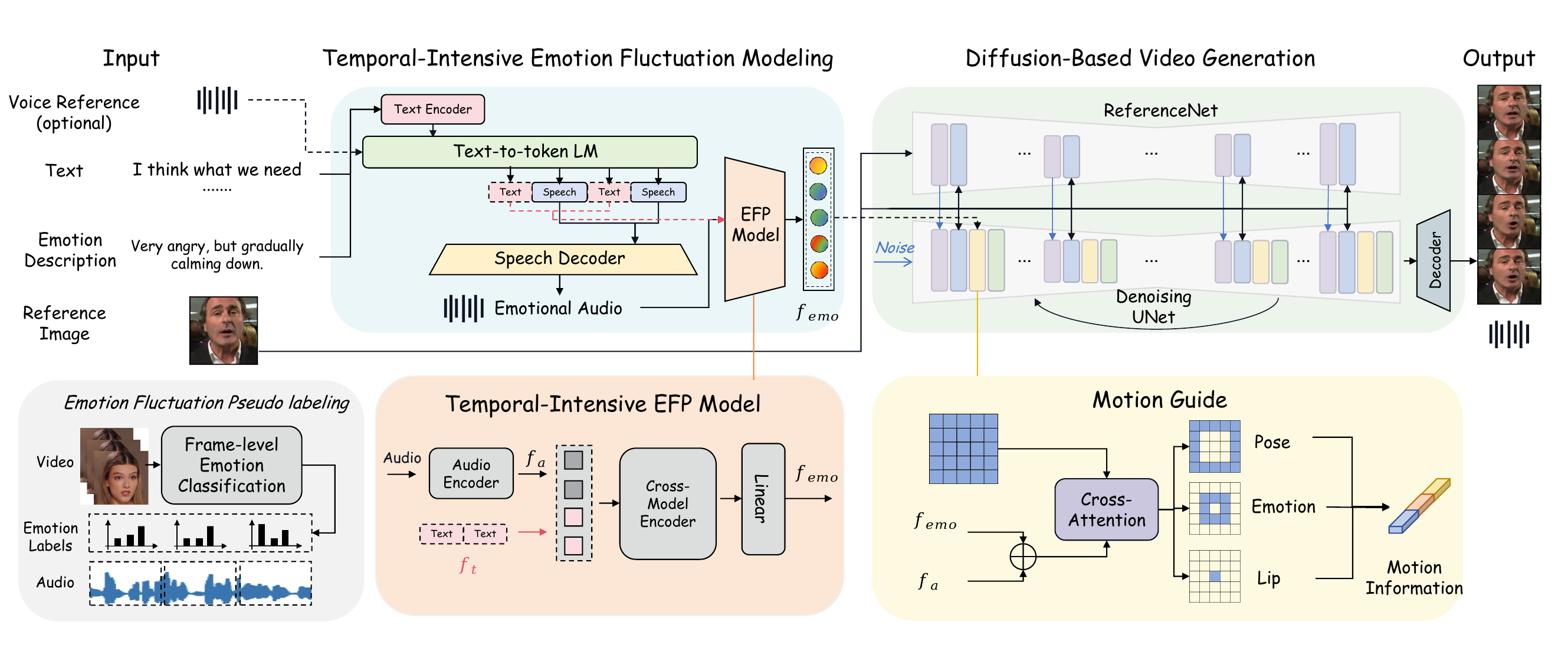}

   \caption{The Overview of TIE-TFG model. Specifically, our framework first utilizes a large-scale TTS model to generate emotional audio based on the text content and emotion description. Next, the emotion fluctuation prediction (EFP) module infers the emotional fluctuations from the audio and text input, with frame-level facial emotion labels serving as its training targets. Finally, we incorporate the obtained emotional fluctuation features into the talking face generation model via a cross-attention mechanism to produce a video with continuous emotional expression.} %图上的temporal-intensive }
   \label{fig1}

\end{figure*}

\subsection{Overview}
Text-driven emotionally continuous and natural talking face generation (EC-TFG) aims to synthesize speaker videos that align with a given emotion description and exhibit human-like emotional dynamics. This task differs from prior audio-driven emotional TFG in three key aspects. First, the driving input shifts from audio to text, enhancing controllability and editability. Second, emotional customization evolves from fixed emotion labels and intensity levels to free-form emotion descriptions, enabling finer-grained emotional control. Third, the generated video must not only reflect the specified emotion but also exhibit continuous emotional fluctuations influenced by the spoken content, aligning more closely with natural human expressions.

The goal of EC-TFG is to generate an emotional talking face video $V$ and corresponding audio $A$ conditioned on input text $T$, emotion description $T_{emo}$, character reference image $I$, and optionally, a voice reference feature $f_{voice}$. In the resulting video and audio, the character’s lip, facial, and head movements are synchronized with the speech and consistent with the intended emotional state.

As illustrated in Figure~\ref{fig1}, we propose TIE-TFG, a model that integrates an emotional TTS module, a temporal-intensive emotion fluctuation prediction model, and an emotion-guided visual synthesis module. We adopt Hallo~\cite{xu2024hallo} as the backbone framework, which employs a stable diffusion-based visual generation model to produce videos. The generation process is further enhanced by ReferenceNet, which leverages reference images to guide visual consistency. Emotional fluctuation features are incorporated during video synthesis to disentangle audio and motion cues in the latent space, enabling fine-grained emotional control. This unified framework supports precise and expressive emotional modulation through natural language descriptions.

\subsection{Emotional Audio Generation}
To produce a talking face that exhibits emotional fluctuations aligned with a specific emotion description, we initially employ a large-scale Text-to-Speech (TTS) model GLM-4-Voice~\cite{zeng2024glm} equipped with emotional customization capabilities. 

\begin{gather}
    A = Emotional\_TTS(T, T_{emo}, f_{voice}),
\end{gather}

The model generates audio that accurately reflects the designated emotion. We extract the intermediate textual representation $f_t$ from the model and use an audio encoder to obtain the audio feature $f_a$.
Following this, we employ a pre-trained emotion fluctuation prediction model to transform the audio and text sequence into a continuous feature sequence representing emotional changes over time. This feature captures variations in both emotion type and intensity. The resulting audio and emotional features serve as motion guidance for a talking face generation model, which produces a video that aligns with the specified emotional dynamics.

\subsection{Temporal-Intensive Emotion Fluctuation Modeling}
To capture the temporal-Intensive  continuous emotional fluctuations in emotional audio, it is necessary to apply detailed emotional labeling to the audio sequence. However, manually labeling such a vast dataset is impractical, leading us to utilize a pseudo-labeling approach. We employ the cutting-edge facial expression emotion prediction model, ResEmoteNet~\cite{roy2024resemotenet}, to determine the emotion and its intensity for each video frame. These predictions serve as surrogate emotional labels for the audio, allowing for precise frame-level emotional annotation. Building on this, to predict the emotion fluctuation sequence, we integrated intermediate textual representations and audio features within the multimodal encoder. The audio features are derived from an enhanced speech emotion representation model, Emotion2vec~\cite{ma2023emotion2vec}, which is extended by incorporating a linear layer and a token-level classification head. The training loss is formulated as follows:
\begin{gather}
        L_{P}(\theta) = -\sum_{i=1}^{N}\textnormal{log}p_\theta(L_{i}|f_{a},f_{t}),
\end{gather}
where $\theta$ denote the model parameters, $N$ represents the sequence length, $L_{i}$ represents the emotion label of the current position.
This modification enabled the training of an audio emotion fluctuation prediction model capable of identifying the sequence features of emotional fluctuations from any audio input.

\subsection{Emotion Fluctuation Guided Visual Synthesis}
% In this section, we will explore the components of the temporal-intensive  Emotion Modulated Talking Face Generation model, TIE-TFG, as depicted in Figure~\ref{fig1}.

After generating emotional audio using a large-scale TTS model based on the text content and emotion description, and inferring the underlying sequence of emotional fluctuations, we use this information as input to the TFG model to generate videos that exhibit corresponding emotional dynamics. In this subsection, we will explore the components of the video generation model, as depicted in Figure~\ref{fig1}.

\textbf{Diffusion Backbone.} This framework leverages a diffusion-based generative architecture with hierarchical audio-visual cross-attention to improve video generation quality. At its core, the model is built upon Stable Diffusion 1.5~\cite{rombach2022high}, a latent diffusion framework consisting of three main components: a Variational Autoencoder (VAE), a U-Net-based denoising network, and a conditioning module. The VAE encodes input images into a continuous latent space and decodes the generated latent representations back into the pixel space, enabling computationally efficient diffusion modeling. Unlike standard text-driven generation tasks, our model uses audio signals and emotional features as the primary motion drivers. Given an image $I\in \mathbb{R}^{H \times W \times 3}$ and its associated conditions $C_{embed}\in \mathbb{R}^{D}$, the latent representation:
\begin{gather}
    z_{0} = \mathcal{E}(I) \in \mathbb{R}^{H_{z} \times W_{z} \times D_{z}}
\end{gather}
Undergoes a diffusion process over T time steps. This process is modeled as a deterministic Gaussian progression, ultimately reaching $z_T \sim \mathcal{N}(0, I)$. The training objective of Stable Diffusion is defined by the following loss function:
\begin{equation} 
L = \mathbb{E}_{\mathcal{E}(I), c_{\text{embed}},\epsilon\sim\mathcal{N}(0,1),t}\left[\lVert \epsilon-\epsilon_{\theta}(z_t, t, c_{\text{embed}}) \rVert_{2}^{2} \right], %=1,...,T
\end{equation} 
In this setup, $t$ is uniformly sampled from $\{1,...,T\}$. Here, $\epsilon_{\theta}$ denotes the model's trainable components, comprising a denoising U-Net with residual blocks and layers dedicated to self-attention and cross-attention. These elements process the noisy latent variable $z_t$ together with the conditional embedding $c_{\text{embed}}$.

Upon training, a deterministic sampling method, such as Denoising Diffusion Implicit Models~\cite{song2020denoising} (DDIM), is employed to reconstruct the initial latent $z_t$. The decoder $\mathcal{D}$ then converts the latent $z_t$ back into the final image output. 

\textbf{ReferenceNet.} To further refine visual fidelity, the framework incorporates ReferenceNet, which guides the generation process using reference images. ReferenceNet shares the Unet layer structure of the denoising network, extracting and aligning features from specific layers to ensure cohesive visual outputs. By referencing pre-existing images, ReferenceNet significantly enhances video quality, especially in terms of portrait and background textures.

\textbf{Motion Guide.} Moreover, a hierarchical visual synthesis module accurately maps audio signals and emotional variations onto visual elements like lip movements, facial expressions, and head pose. 

Using the MediaPipe\footnote{https://github.com/google-ai-edge/mediapipe} toolbox, we extract landmarks from the face image $I$, generating masks $M_{\text{lip}}, M_{\text{exp}}, M_{\text{pose}} \in\{0,1\}^{H_z\times W_z}$, which represent the lip, expression, and pose regions, respectively.
To combine the audio features $f_{a}$ and the emotional fluctuation features $f_{emo}$, we utilize a primary-guided weighted fusion strategy, with the weight $g(x)$ dynamically adjusting based on the audio features.
\begin{gather}
    c_{\text{fusion}} = f_{a} + g \cdot f_{emo}, \\
    g = \sigma(W_{g} \cdot f_{a} + b_{g}),
\end{gather}
Here, $\sigma(\cdot)$ denotes the sigmoid function, which projects the gate value into the range $(0,1)$. $W_g$ and $b_g$ are learnable parameters, where $W_g$ is the weight matrix and $b_g$ is the bias term.

Following this, we incorporate a cross-attention mechanism between the latent representations and the fused embeddings.
\begin{gather}
    o_t = \text{CrossAttn}(z_t, c_{\text{fusion}}).
\end{gather}
Masks are then applied to produce multi-scale latent representations according to this approach.
\begin{gather}
 b_t = o_t\odot M_{\text{pose}}; 
 f_t = o_t\odot M_{\text{exp}};
 l_t = o_t\odot M_{\text{lip}}.
\end{gather}
Finally, an adaptive module is introduced to manage hierarchical audio guidance.
\begin{gather}
    \sum_u \text{Conv}_u(u), \; u\in\{a_t, f_t, l_t\}.
\end{gather}
This temporal alignment, achieved through multiple self-attention blocks along the temporal axis, ensures consistent and coherent transitions across video frames.

\subsection{Training and Inference}

During the training phase, we only train the independent emotion fluctuation prediction model and the motion alignment component of the emotion-guided TFG model. The emotion fluctuation prediction model is trained as described in Section 3.2. For the emotion-guided TFG model, we keep ReferenceNet and Denoising UNet fixed, and we train the video sequences using reference images, input audio, emotion fluctuation features, and target video data. The focus is on enhancing the ability of the video sequence generation under the guidance of emotion fluctuation features. This phase primarily emphasizes training hierarchical audiovisual cross-attention to establish relationships between audio, emotion fluctuation, and the visual information of lips, expressions, and poses. During this phase, a random frame from the video clip is selected as the reference image.

In the inference phase, the network takes a single reference image, driving text, emotion description, and optionally tone information as input. It first generates audio that matches the emotion description and predicts emotion fluctuation features, then generates a video sequence of the animated reference image based on the corresponding audio and emotion fluctuation features.

\section{Experiment}

\begin{table*}[!t]
\centering
\footnotesize
\resizebox{0.95\textwidth}{!}{
\begin{tabular}{c|c|c|c|cc|cc|c|c|c}
\toprule
Method & Dataset  &Input & Audio   & \multicolumn{1}{c}{FID ↓} & \multicolumn{1}{c|}{FVD ↓} & \multicolumn{1}{c}{PSNR ↑} & \multicolumn{1}{c}{SSIM ↑} & \multicolumn{1}{c|}{Sync-D ↓} & \multicolumn{1}{c|}{E-FID ↓} & \multicolumn{1}{c}{EF-score ↑} \\ 
\midrule
\multirow{4}{*}{}
MakeItTalk\cite{zhou2020makelttalk} &  & V+A & TTS   & 25.168 & 245.786 & 17.25  & 0.5562 & 10.748 & 15.624 & 32.56 \\
TTFS*\cite{jang2024faces} &  & V+T & -   & 19.620 & - & 30.28  & 0.9436 & - & - & - \\
FT2TF*\cite{diao2025ft2tf} & LRS2 & V+T & -   & 22.520 & - & 33.20 & 0.9901  & - & - & - \\
\rowcolor{mygray}
TIE-TFG (Ours) &  & V+T & TTS  & 18.763 & 210.420 & 36.96 & 0.9910 & 7.486 & 8.153 & 75.84 \\
\rowcolor{mygray}
TIE-TFG (Ours) &  & V+T & GT   & \textbf{17.942} & \textbf{204.386} & \textbf{37.20} & \textbf{0.9922} & \textbf{7.245} & \textbf{8.097} & \textbf{78.65} \\
\midrule
\multirow{7}{*}{}

% Audio2Head\cite{wang2021audio2head}  & 37.776 & 239.860 & \textbf{8.024} & \textbf{7.145} & 17.103 & 39.98 \\
DreamTalk\cite{ma2023dreamtalk} &  & V+A & GT    & 78.147 & 790.660 & 14.30 & 0.7848 & 8.364 & 15.696 & 32.21 \\
AniPortrait\cite{wei2024aniportrait} & & V+A & GT & 26.561 & 234.666 & 17.92 & 0.8329 & 10.548 & 13.754 & 44.80 \\
% Hallo~\cite{xu2024hallo}                                & \textbf{20.545} & \textbf{173.497} & 7.750 & 7.659 & \textbf{7.951} & 50.42 \\
% Hallo*                               & 22.345 & 196.394 & 6.970 & 7.458 & 8.430 & 45.43 \\
SadTalker\cite{zhang2022sadtalker} & & V+A & GT   & 22.340 & 203.860 & 30.14  & 0.9428  & 7.545 & 9.776 & 30.76 \\
Hallo~\cite{xu2024hallo} & HDTF  & V+A  & TTS & 23.328 & 205.924  & 33.68 & 0.9480 & 7.667 & 8.562 & 43.62 \\
Hallo~\cite{xu2024hallo} &   & V+A & GT                            & 22.345 & 196.394 & 34.09 & 0.9550  & 7.458 & 8.430 & 45.43 \\ 
\rowcolor{mygray}

TIE-TFG (Ours) &  & V+T & TTS & 22.103 & 193.800 & 36.85 & 0.9915 & 7.516 & 8.371 & 74.62 \\
\rowcolor{mygray}
TIE-TFG (Ours)   &   & V+T  & GT                           & \textbf{21.362} & \textbf{186.864} & \textbf{37.28} & \textbf{0.9930} & \textbf{7.249} & \textbf{8.298} & \textbf{77.24} \\ 
% \midrule
% Real Video        &   -        & -     &  -     &  -      & 8.700 & 6.597 & -     & 100.00 \\
\bottomrule
\end{tabular}
}
\caption{
Quantitative comparison with existing talking face generation approaches on the HDTF and LRS2 dataset in the one-shot setting. The Audio column refers to the speech source for generation (GT: ground truth, TTS: synthesised audio, -: end-to-end method, no audio used as intermediate representation). * indicates that the data for this method is taken from the original paper.
% Our proposed method is able to generate high-quality, temporally coherent talking head animations with excellent lip sync performance and improved emotional fluctuations. 
}
\label{tab:quantitative_hdtf}
\end{table*}

\begin{table}[!t]
\centering
\footnotesize
\resizebox{.4\textwidth}{!}{
\begin{tabular}{c|c|cc}
\toprule
Method & \multicolumn{1}{c|}{FID ↓} & \multicolumn{1}{c}{Emo-Acc ↑} & \multicolumn{1}{c}{EF-score ↑} \\ 
\midrule
Wav2Lip~\cite{schneider2019wav2vec} & 67.49 & 17.87 & 10.34 \\
EAMM~\cite{ji2022eamm} & 22.38 & 49.85 & 32.55  \\
EAT~\cite{gan2023efficient} & 19.69 & 75.43 & 47.28 \\
% MM-ESL~\cite{xu2023high} & 15.89 & 80.34 & 60.48 \\
% SPACE~\cite{gururani2023space} &16.67 & 81.32 & 62.76 \\
FWEI*~\cite{xu2024learning} & 16.77 & 83.55 & - \\
\rowcolor{mygray}
TIE-TFG                                  & \textbf{15.27} & \textbf{84.05} & \textbf{67.58}  \\
\midrule
Real video                           &  0     &  84.37      & 100.00  \\
\bottomrule
\end{tabular}
}
\caption{
The quantitative comparisons with the existing TFG approaches on the MEAD dataset. * indicates that the data for this method is taken from the original paper.
% Our method is able to generate high-quality, temporally coherent talking head animations with excellent lip sync performance and improved emotional fluctuations.
}
\label{tab:quantitative_mead}
\end{table}

\subsection{Experimental Setups}

\paragraph{Dataset.} We utilized the Voxceleb2~\cite{chung2018voxceleb2}, LRS2~\cite{afouras2018deep}, HDTF~\cite{zhang2021flow}, MEAD~\cite{wang2020mead}, and CREMA-D~\cite{cao2014crema} datasets to train the temporal-intensive emotion fluctuation model and the talking face generation model. We selected videos totaling 10 hours to create the EC-HDTF dataset for training the emotion-continuous talking face generation model. For predicting emotional fluctuations, we employed the emotion classification model ResEmoteNet~\cite{roy2024resemotenet} to analyze facial expressions in each video frame, generating pseudo-labels for emotional fluctuations, and divided the dataset into an 8:2 split for training and testing. Furthermore, to validate the effectiveness of our approach, we compare with one-shot talking face generation methods on the MEAD test set. Throughout the testing of the overall framework, we leveraged the iFlytek\footnote{https://global.xfyun.cn/products/lfasr} speech-to-text model and the GLM-4-Voice~\cite{zeng2024glm} text-to-speech model for audio and text conversion, and manually identify and annotate emotion descriptions.

\paragraph{Evaluation Metrics.}
The evaluation metrics include Fréchet Inception Distance (FID), Fréchet Video Distance (FVD), Peak Signal-to-Noise Ratio (PSNR), Structural Similarity (SSIM), Synchronization-D (Sync-D), and E-FID. Specifically, FID and FVD assess the similarity between generated images and real data, with lower values indicating better performance and more realistic outputs. Sync-D evaluate the lip synchronization quality of generated videos. E-FID measures the quality of generated images using features extracted from the inception network, providing a more nuanced assessment of fidelity.

Furthermore, to assess the accuracy of the generated emotions and facilitate comparison with existing works, we follow the approach in~\citet{gan2023efficient}, fine-tuning emotion classification model on the MEAD training set for measure emotion accuracy, referred to as the Emo-Acc metric. To assess how effectively the emotional fluctuations in the video are modeled, we introduce the Emotional Fluctuation Score (EF-score) metric. This metric evaluates the consistency of emotion labels between the original and generated videos on a frame-by-frame basis. We measure label consistency using accuracy, defining the EF-score as the proportion of frames where the emotion labels match across the dataset. 

Compared with EMO-Acc, which recognizes emotions for the entire video, EF-score focuses on the emotion accuracy at the frame level. Its calculation is shown as follows:
\begin{equation}
\text{EF-score} = \frac{1}{N} \sum_{i=1}^{N} \mathbb{I} \left( \mathcal{F}(v^{(i)}_{\text{orig}}) = \mathcal{F}(v^{(i)}_{\text{gen}}) \right),
\end{equation}
where $N$ denotes the total number of frames in the evaluation set, $v^{(i)}_{\text{orig}}$ and $v^{(i)}_{\text{gen}}$ are the $i$-th frames of the original and generated videos, respectively, and $\mathbb{I}(\cdot)$ is the indicator function that returns 1 if the predicted emotion labels match, and 0 otherwise. To avoid potential evaluation bias, we use an independently trained ResEmoteNet model as $\mathcal{F}$, ensuring that the model used for assessment is distinct from the one used for generating pseudo labels during training. For details on the comparative methods and additional experimental design, please refer to Appendix.

% Considering that the model used for emotion fluctuation annotation should be distinct from the one employed in testing, we specifically train an additional ResEmoteNet model. We ensure that the testing model possesses equivalent capability while avoiding potential biases introduced by using the same model for both annotation and assessment.

\subsection{Quantitative Results}
Table~\ref{tab:quantitative_hdtf} provides a comprehensive quantitative evaluation on the HDTF and LRS2 dataset compared to existing TFG techniques. The results indicate that our method significantly outperforms previous approaches in terms of EF-score, thanks to the effective modeling of emotional fluctuations and visual motions. Our approach achieves lower FID, FVD, and E-FID scores than most of the baseline, while exhibits a high degree of consistency with real videos in lip synchronization metrics. These findings underscore the importance of emotional fluctuation modeling in generating high-quality and temporally coherent talking face videos. At the same time, we can observe that compared to using ground truth (GT) audio, the video generation performance deteriorates when guided by audio generated from an emotional TTS system. This indicates that the modeling of emotional fluctuations is limited by the performance of existing emotional TTS models. Therefore, in our framework, we incorporate textual features into the emotional fluctuation prediction process to enhance the emotional dynamics in the generated videos. We also conducted ablation studies on both the textual and audio features in the Table~\ref{tab:efp}.

Table~\ref{tab:quantitative_mead} presents the results on the MEAD test set, demonstrating that our model outperforms others across most metrics, especially in Emo-Acc (83.92) and EF-score (66.45). These results highlight the superiority of the emotional expressions captured by our proposed method.

\begin{figure}

\begin{center}

\includegraphics[width=7.5cm]{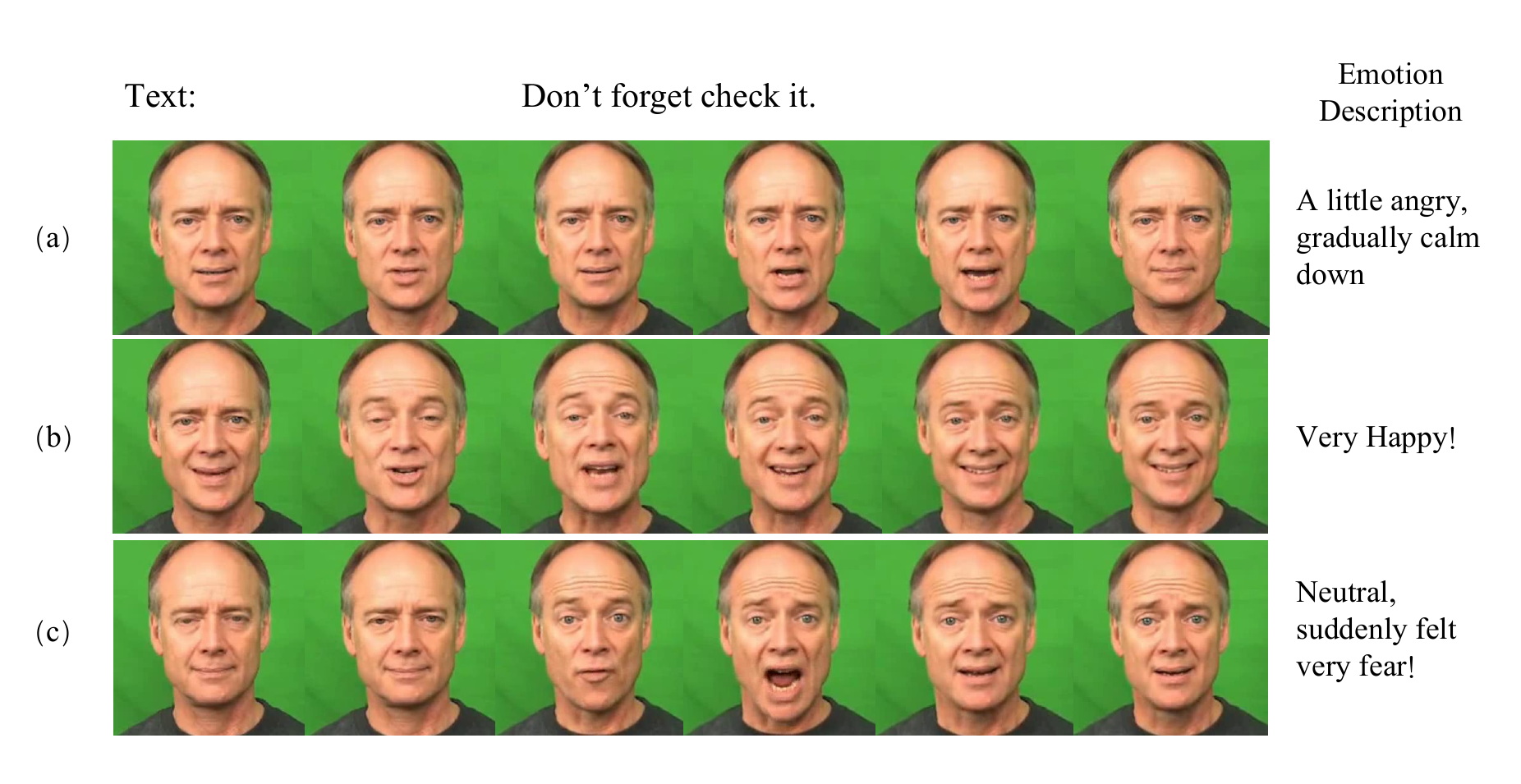} 

\caption{
Video generation results of the proposed approach given different emotion description.
}
\label{fig:case}
\end{center}

\end{figure}

\subsection{Qualitative Results}
\begin{figure}[!ht]

\begin{center}

\includegraphics[width=6.6cm]{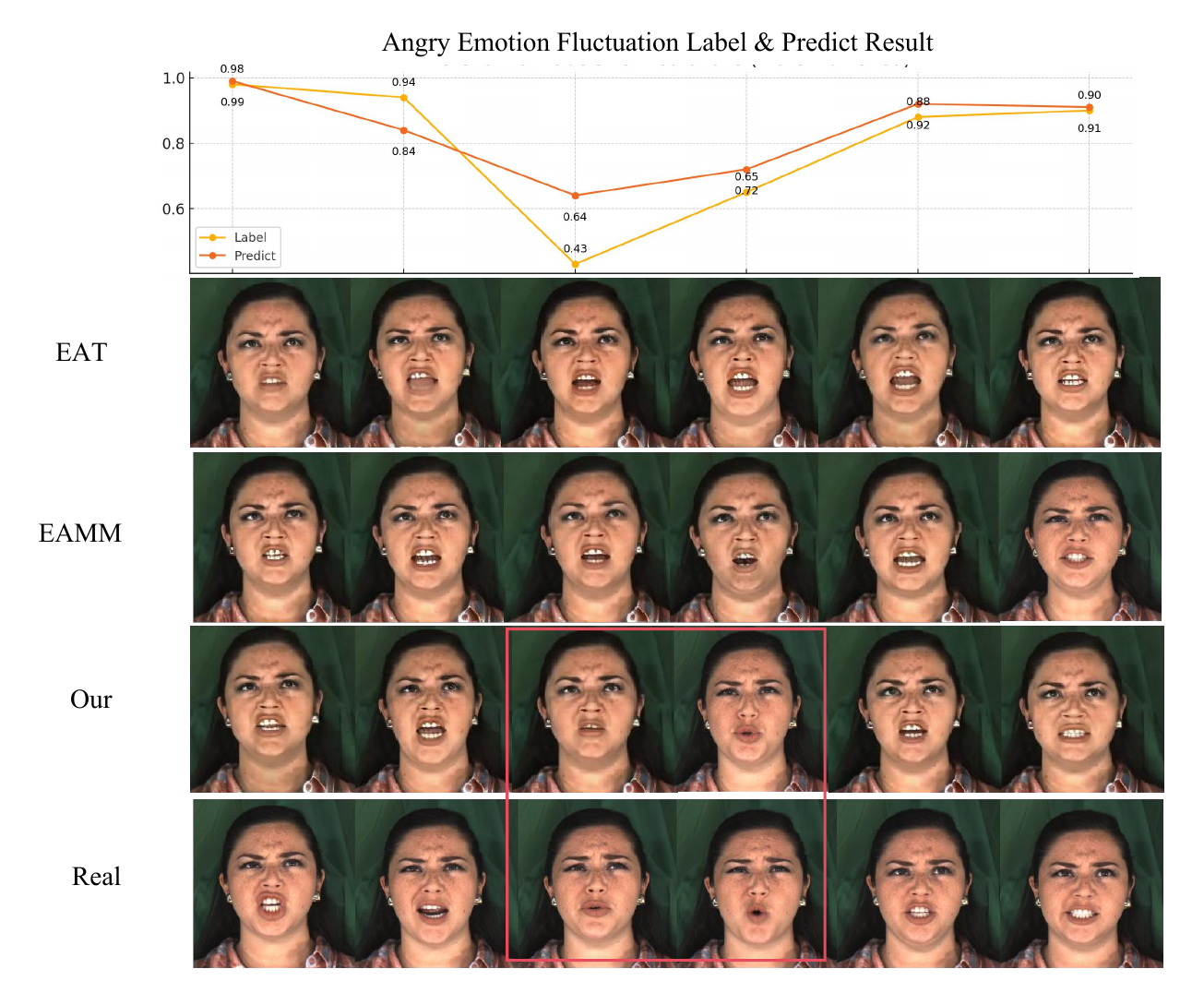} 
\caption{
Comparison of our method with existing emotion control approaches. The top part illustrates the predicted emotional fluctuations versus the pseudo-emotional labels from the reference video. The bottom part displays the generated results from different emotion control methods.
% We compare our approach with several emotion control methods for audio-driven talking face generation. Both the emotion label and the description are set to ``angry''. The top part of the figure presents a comparison between the predicted emotional fluctuations and the pseudo-emotional labels of the reference video, while the bottom part shows the outputs of different emotion control methods.
% We compared our approach with several emotion control methods used for audio-driven talking face generation. The emotion label and emotional description were set to ``angry'' at the top, we present a comparison between the predicted emotional fluctuations and the pseudo-emotional labels of the real video. The bottom shows the results of different emotion control methods. 
% It can be observed that our method produces results more similar to the real video, including a certain degree of emotional fluctuation, as indicated by the red-bounded box, resulting in more diverse and realistic talking face outputs. In contrast, other methods tend to generate results with a fixed emotion.
}
\label{fig:ef}
\end{center}

\end{figure}

In Figure~\ref{fig:ef}, we provide a visual comparison between our method and existing emotion control techniques. At the top are the predictions from the audio emotion fluctuation predictor, along with pseudo-emotion labels generated through frame-wise facial analysis, which are used to create an angry talking face video. The first and second rows depict the results from EAT~\cite{gan2023efficient} and EAMM~\cite{ji2022eamm}, both using constant emotion intensity, where facial expressions remain uniformly angry with minimal variation. It can be observed that our method produces results more similar to the real video, including a certain degree of emotional fluctuation, as indicated by the red-bounded box, resulting in more diverse and realistic talking face outputs. In contrast, other methods tend to generate results with a fixed emotion. 
% In contrast, our proposed method yields more diverse and realistic results by accurately controlling emotional fluctuation features, resulting in more natural and lifelike talking face videos.

\subsection{Analysis on Emotion Controllability}
Figure~\ref{fig:case} illustrates the emotional control capabilities of our proposed method. Given the same text input with varying emotion descriptions, our method generates corresponding videos and audio that align with these descriptions, achieving synchronized emotion-motion dynamics. (a) and (c) display the generated results for different emotional variations, highlighting the ability of our approach to provide fine-grained emotion control through precise descriptions. (b) presents the results for fixed emotion and intensity, showing that our method can adapt speech rhythm according to the text content while conveying emotions, thereby enabling more natural emotional control with adaptive intensity variations.

\subsection{EC-TFG Quantitative Evaluation of Generated Audio Quality}

To verify the acceptability of the emotional audio generated by the TTS model(GLM-4-Voice), we conducted a quantitative evaluation by comparing 100 randomly selected real audio samples (GT) from CREMA-D and MEAD with the generated audio. The text content for the generated audio was obtained from ASR transcriptions of the real audio. MEAD and CREMA-D datasets contain a large number of emotional audio clips with identical textual content, making it ideal for verifying the TTS model's ability to generate emotional audio under different emotion descriptions. Specifically, we focused on the fluency and emotional accuracy of the generated audio, using the following metrics: a) Fluency: measured by Word Error Rate (WER).
b) Emotional Accuracy: evaluated using an external audio emotion classifier (Wave2Vec) by comparing the predicted emotion labels with those of the GT.
c) Subjective Evaluation: 10 human evaluators rated the audio pairs based on naturalness and emotional expressiveness (using a Likert scale).

\begin{table}[!ht]
\centering
\scalebox{0.8}{
\begin{tabular}{lcc}
\toprule
\textbf{Metric} & \textbf{MEAD} & \textbf{CREAD} \\
\midrule
WER (Synthesized vs.\ GT)           & 5.1\%   & 6.3\%   \\
Emotion Accuracy (Match with GT)     & 93.2\%  & 95.4\%  \\
Subjective Rating (Naturalness / 5)  & 4.2     & 4.0     \\
\bottomrule
\end{tabular}}
\caption{Comparison of Synthesized vs.\ Ground Truth Audio.}
\label{tab:comparison}
\end{table}

Table~\ref{tab:comparison} presents the results, confirming that the TTS model we employed produces outputs with acceptable fluency and emotional expressiveness compared to the GT. This supports the subsequent process of our framework.

\subsection{Ablation Study on Emotion Description Complexity}

To validate the performance of our proposed temporal-intensive  emotion fluctuation prediction module on audio generated under complex emotion descriptions, we compared the emotion label prediction accuracy between audio generated with a single emotion label and audio with two or more emotion label variations. We manually selected 20 videos for each condition and used an independent frame-level emotion prediction model for annotation, with the annotations manually verified. The results are presented in Table \ref{tab:emotion_performance}.

\begin{table}[htbp]
    \centering
    \scalebox{0.7}{
    \begin{tabular}{lcc}
        \toprule
         & Accuracy  & F1-Score \\
        \midrule
        Single-label            & \textbf{67.88}                & \textbf{66.90}                \\
        Multi-label               & 66.45                & 65.72                \\
        \bottomrule
    \end{tabular}}
    \caption{Comparison of temporal-intensive emotion fluctuation prediction (EFP) model performance on audio with single and multiple emotion labels.}
    \label{tab:emotion_performance}
\end{table}

Compared to single-label audios, the module's performance on multi-label audios does not significantly decline. This demonstrates that the module has the ability to predict the emotional fluctuations in audio generated from complex emotion descriptions.

\subsection{Ablation of Emotional Fluctuation Features}
To assess the impact of inaccuracies in the emotion fluctuation prediction module, we conducted an ablation study by replacing the predicted features with random noise. As shown in the table below, while performance slightly degrades, the decline is limited due to the gating mechanism that mitigates noise effects. These results confirm the positive contribution of emotion fluctuation features while demonstrating the model’s robustness to prediction noise. 

\begin{table}[ht]
  \centering
  \scalebox{0.6}{
  \begin{tabular}{lccccc}
    \toprule
    Method & FID $\downarrow$ & FVD $\downarrow$  & Sync-D $\downarrow$ & E-FID $\downarrow$ & EF-score $\uparrow$ \\
    \midrule
    
    % Hallo  & 22.345 & 196.394 & 6.970 & 7.458 & 8.430 & 45.43 \\
    Our    & \textbf{21.362} & \textbf{186.864}  & \textbf{7.249} & \textbf{8.298} & \textbf{77.24} \\
    % Our    & \textbf{21.854} & \textbf{187.422} & \textbf{7.032} & \textbf{7.398} & \textbf{8.359} & \textbf{76.34} \\
    w/o EF  & 22.345 & 196.394  & 7.458 & 8.430 & 45.43 \\
    w/ Random & 24.512 & 208.376  & 7.843 & 9.265 & 39.62 \\
    \bottomrule
  \end{tabular}}
  \caption{Ablation of emotional fluctuation features.}
  \label{tab:metrics_comparison}
\end{table}

\subsection{Analysis on Emotion Fluctuation Pseudo-labeling}
Modeling emotional fluctuations requires frame-level emotion labels to align audio with visual expressions. Notably, studies~\cite{cao2014crema} indicate that recognition accuracy for manually labeled target emotions is 40.9\% for audio-only, 58.2\% for video-only, and 63.6\% for combined video and audio modalities. 
Given the low accuracy and high cost of manually annotating frame-level emotional fluctuations compared to existing emotion recognition models, we opt to use state-of-the-art visual emotion recognition results as pseudo-labels. This approach allows us to bypass the need for manual annotation when generating emotion fluctuation sequences.
% Given that manually identifying frame-level emotional fluctuations is less accurate than recognizing coarse-grained emotions and is also costly, we aim to avoid manual annotation for obtaining emotion fluctuation sequences.

To generate pseudo-labels, we utilize the state-of-the-art model ResEmoteNet~\cite{roy2024resemotenet} for the emotion classification task. 

\subsection{Ablation of Emotion Fluctuation Prediction Model}
Since accurate prediction of emotional fluctuations is equally crucial, we compare existing audio representation methods for emotional fluctuation modeling in Table and select the best-performing Emotion2vec~\cite{ma2023emotion2vec} model. Meanwhile, Table~\ref{tab:efp} presents an ablation study on different input modalities. The results show that combining audio and text information yields the best prediction performance.

% To generate pseudo-labels, we utilize the state-of-the-art model ResEmoteNet~\cite{roy2024resemotenet}
% % , which achieves an accuracy of 100\% on CK+~\cite{lucey2010extended} and 79.79\% on FER2013~\cite{goodfellow2013challenges} 
% for emotion classification task. Accurate emotional fluctuation prediction is equally crucial; we compare existing audio representation methods for emotional fluctuation modeling in Table~\ref{tab:efp} and select the best-performing Emotion2vec model.

\begin{table}[!t]
\centering
\footnotesize
\resizebox{.47\textwidth}{!}{
\begin{tabular}{l|cccc}
\toprule
Method & \multicolumn{1}{c}{Accuracy} & \multicolumn{1}{c}{Precision} & \multicolumn{1}{c}{Recall} & \multicolumn{1}{c}{F1-score}  \\ 
\midrule
% Random & 12.70 & 22.89 & 12.70 & 15.29 \\
Wav2Vec~\cite{schneider2019wav2vec} & 40.26 & 40.43 & 38.13 & 39.26 \\
HuBERT~\cite{hsu2021hubert} & 62.93 & 62.39 & 62.94 &62.53 \\
% \midrule
\rowcolor{mygray}
Our & \textbf{68.28} & \textbf{67.59} & \textbf{68.33} & \textbf{67.98}\\
- w/o Text & 65.34 & 64.78 & 66.12 & 65.45 \\
- w/o Audio & 63.89 & 63.45 & 64.75 & 64.11 \\
% Emotion2vec~\cite{ma2023emotion2vec} & \textbf{67.63} & \textbf{66.92} & \textbf{67.64} & \textbf{66.84}\\
\bottomrule
\end{tabular}
}
\caption{
Ablation study of input features and audio backbone used in the emotion fluctuation prediction model.
% Comparison of different audio-based emotion classification models for emotional fluctuation prediction.
}
\label{tab:efp}
\end{table}

In addition, we supplemented the quantitative evaluation of the image emotion classification model, the human evaluation and emotion conflict analysis in the appendix.

\section{Conclusion}
Conventional audio-driven talking face generation (TFG) methods can only convey fixed emotions embedded in audio, limiting their flexibility in real-world applications. To address this limitation, we propose the first text-driven TFG framework that models emotion fluctuations to enable fine-grained and dynamic emotional control. Our approach incorporates an emotion fluctuation predictor to capture temporal changes in emotions and leverages these features to decouple audio and motion information in the latent space, thereby guiding emotion-controllable facial synthesis. Experimental results demonstrate that our framework achieves high-quality, customizable, and expressive talking face generation, significantly advancing the state of the art.

% \section{Acknowledgments}
% AAAI is especially grateful to Peter Patel Schneider for his work in implementing the original aaai.sty file, liberally using the ideas of other style hackers, including Barbara Beeton. We also acknowledge with thanks the work of George Ferguson for his guide to using the style and BibTeX files --- which has been incorporated into this document --- and Hans Guesgen, who provided several timely modifications, as well as the many others who have, from time to time, sent in suggestions on improvements to the AAAI style. We are especially grateful to Francisco Cruz, Marc Pujol-Gonzalez, and Mico Loretan for the improvements to the Bib\TeX{} and \LaTeX{} files made in 2020.

% The preparation of the \LaTeX{} and Bib\TeX{} files that implement these instructions was supported by Schlumberger Palo Alto Research, AT\&T Bell Laboratories, Morgan Kaufmann Publishers, The Live Oak Press, LLC, and AAAI Press. Bibliography style changes were added by Sunil Issar. \verb+\+pubnote was added by J. Scott Penberthy. George Ferguson added support for printing the AAAI copyright slug. Additional changes to aaai2026.sty and aaai2026.bst have been made by Francisco Cruz and Marc Pujol-Gonzalez.

% \bigskip
% \noindent Thank you for reading these instructions carefully. We look forward to receiving your electronic files!

\bibliography{aaai2026}

\appendix

\newpage
\section{Experiment Details}

\paragraph{Implementation Details.}
% Experiments, including both training and inference, were carried out on a computing platform equipped with 8 NVIDIA A100 GPUs.
All experiments, including training and inference, were conducted on a platform with 8 NVIDIA A100 GPUs.
For emotion-aware generation, both the initial and subsequent phases consisted of 30,000 training steps, with a batch size of 4 and a video resolution of $512 \times 512$. Each training instance generated 14 video frames, with the latent frames from the motion module concatenated to the first two ground truth frames to ensure video continuity. A learning rate of 1e-5 was used during training, with the model initialized using weights from Hallo. To improve video generation quality, reference images, guiding audio, and motion frames were dropped with a probability of 0.05 throughout training. For the emotional fluctuation prediction model, a learning rate of 1e-3 was used, and the model was initialized with weights from emotion2vec.

\paragraph{Comparison Method.} To evaluate the effectiveness of our proposed approach, we conduct both quantitative and qualitative comparisons between TIE-TFG and several representative baseline methods. These baselines include one-shot, audio-driven, high-quality talking face generation approaches such as SadTalker~\cite{zhang2023sadtalker}, DreamTalk~\cite{ma2023dreamtalk}, AniPortrait~\cite{wei2024aniportrait}, and Hallo~\cite{xu2024hallo}, as well as methods that support emotion editing, including Wav2Lip~\cite{schneider2019wav2vec}, EAMM~\cite{ji2022eamm}, EAT~\cite{wang2024eat}, 
% MM-ESL~\cite{xu2023high}, SPACE~\cite{gururani2023space}, 
and FWEI~\cite{xu2024learning}.

\section{Quantitative Evaluation of the Image Emotion Classification Model}
Since our method for predicting emotional fluctuations relies on the performance of a frame-level image emotion classification model, we further evaluated this model through manual annotation and quantitative analysis on the HDTF and CREMA-D datasets.

\begin{table}[htbp]
  \centering
  \scalebox{0.7}{
  \begin{tabular}{lccccc}
    \hline
    Dataset   & Frames & Accuracy  & Precision  & Recall & F1-score \\
    \hline
    HDTF      & 512             & 87.0          & 85.0                 & 84.0              & 84.4      \\
    CREMA-D   & 1,024             & 85.3          & 84.5                 & 83.8              & 84.1        \\
    \hline
  \end{tabular}}
  \caption{Overall performance of the image emotion classification model on the HDTF and CREMA-D datasets.}
  \label{tab:overall_metrics}
\end{table}

\begin{table}[htbp]
  \centering
  \scalebox{0.78}{
  \begin{tabular}{lccccc}
    \hline
    Emotion  & Accuracy  & Precision  & Recall  & F1-score  \\
    \hline
    Happy    & \textbf{90.5}          & \textbf{90.1}           & \textbf{89.0}        & \textbf{89.5}    \\
    Angry    & 88.0          & 88.3           & 87.5        & 87.9    \\
    Sad      & 85.0          & 85.0           & 84.2        & 84.6    \\
    Fear     & 83.0          & 83.5           & 82.8        & 83.1    \\
    Surprise & 86.5          & 86.7           & 85.9        & 86.3    \\
    Neutral  & 87.5          & 87.8           & 87.1        & 87.4    \\
    \hline
  \end{tabular}}
  \caption{Performance metrics per emotion label on the HDTF dataset and the CREMA-D dataset.}
  \label{tab:emotion_metrics}
\end{table}

We randomly selected video frames from the HDTF and CREMA-D datasets and allowed our model to predict their emotion labels. Subsequently, three human experts evaluated the accuracy of the generated labels using a voting mechanism. Table~\ref{tab:overall_metrics} presents the overall performance of the image emotion classification model, while Table~\ref{tab:emotion_metrics} provides the evaluation results for each emotion label. It can be observed that the frame-level image emotion classification model achieves an overall accuracy exceeding 85\% on both datasets used in this study. Moreover, due to a higher proportion of neutral images in the randomly selected HDTF sample, the model performs better on this dataset compared to the more emotionally diverse CREMA-D dataset. These findings confirm that the model is capable of capturing dynamic emotional fluctuations in videos.

\begin{figure}[!ht]
\begin{center}
\includegraphics[width=8cm]{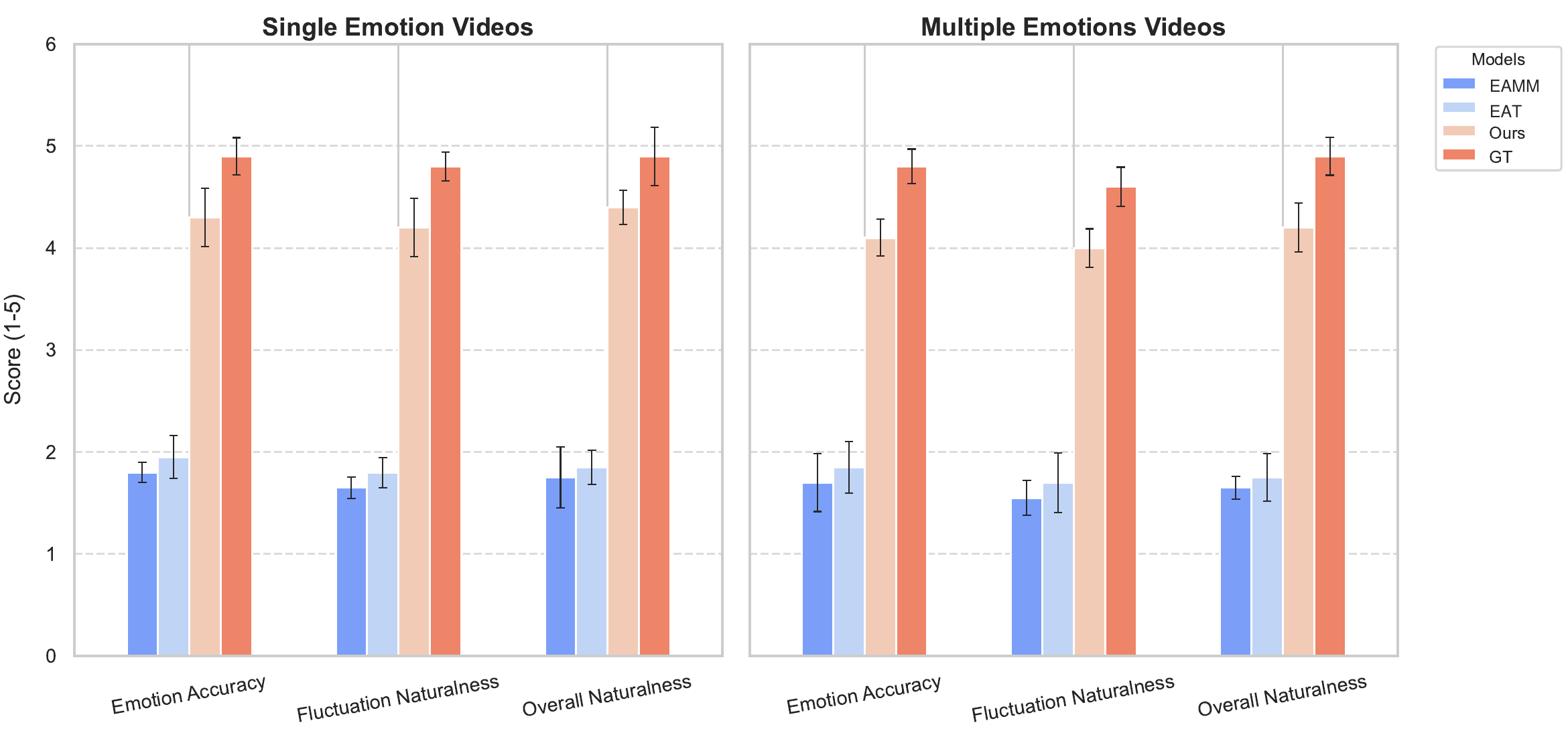} 
\caption{
User studies on the quality of generated talking-face results.
}
\label{fig:us}
\end{center}
\end{figure}
\section{Human Evaluation}
We also conducted a user study involving 20 volunteers (10 male, 10 female) to evaluate the reliability of our method in generating speaker videos. For each method, we generated 20 videos, including 10 videos with a single specified emotion and 10 videos containing multiple emotions. Participants were asked to evaluate the generated results based on three aspects: (1) Accuracy of Emotion Control, (2) Effectiveness of Emotion Fluctuation , and (3) Overall Naturalness.

As illustrated in Figure~\ref{fig:us}, aside from the real data baseline, our method achieved the highest scores in both overall naturalness and accuracy of emotion control under both single-emotion and multi-emotion conditions. Furthermore, compared with other methods, our approach also demonstrated the best performance in terms of the effectiveness of emotion fluctuation. These results indicate that our method aligns more closely with human perception of realistic emotional behavior.

\section{Emotion Conflict Analysis}

\begin{figure}[]

\begin{center}

\includegraphics[width=8cm]{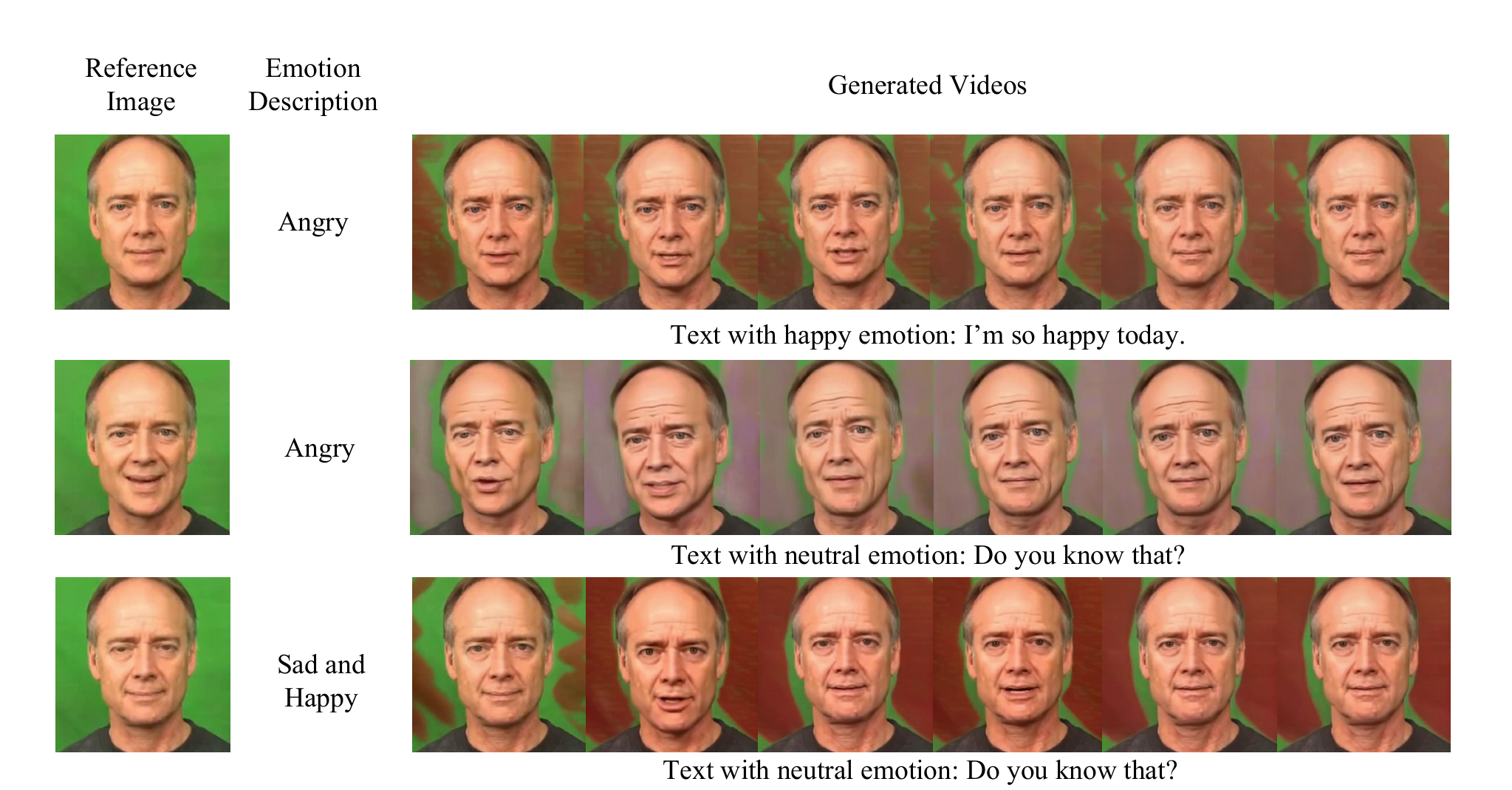}

\caption{
Emotional conflict case analysis. We present several emotional conflict cases of the proposed method in situations where emotional conflicts arise among the input modalities, namely, the emotion description, reference image, and input text.
}
\label{fig:case2}
\end{center}

\end{figure}

While text-driven TFG generation improves editability, the model's performance noticeably declines when dealing with inputs containing conflicting emotions. As illustrated in Figure~\ref{fig:case2}, we evaluated the effects of conflicting emotional control using MEAD on a model that had not been trained with MEAD. In the first row, the given emotion description is ``angry'', while the text conveys a ``happy'' sentiment. In the second row, the reference image shows a smiling face, but the specified emotion is ``angry''. In the third row, the given emotion description includes two incompatible emotions. It can be seen that in the first two cases, despite the conflicting emotional cues, the stable emotion descriptions influence audio generation, enabling the model to produce results consistent with the specified emotions. However, in the third case, with conflicting emotion descriptions, the control fails, suggesting that effective complex emotional control still requires ensuring emotional coherence.

\end{document}